# Efficient Selective Audio Masked Multimodal Bottleneck Transformer for Audio-Video Classification
Wentao Zhu

## Abstract

*Audio and video are two most common modalities in the mainstream media platforms, e.g., YouTube. To learn from multimodal videos effectively, in this work, we propose a novel audio-video recognition approach termed audio-video Transformer, AVT, leveraging the effective spatio-temporal representation by the video Transformer to improve action recognition accuracy. For multimodal fusion, simply concatenating multimodal tokens in a cross-modal Transformer requires large computational and memory resources, instead we reduce the cross-modality complexity through an audio-video bottleneck Transformer. To improve the learning efficiency of multimodal Transformer, we integrate self-supervised objectives, i.e., audio-video contrastive learning, audio-video matching, and masked audio and video learning, into AVT training, which maps diverse audio and video representations into a common multimodal representation space. We further propose a masked audio segment loss to learn semantic audio activities in AVT. Extensive experiments and ablation studies on three public datasets and two in-house datasets consistently demonstrate the effectiveness of the proposed AVT. Specifically, AVT outperforms its previous state-of-the-art counterparts on Kinetics-Sounds by 8%. AVT also surpasses one of the previous state-of-the-art video Transformers [25] by 10% on VGGSound by leveraging the audio signal. Compared to one of previous state-of-the-art multimodal method, MBT [32], AVT is 1.3× more efficient in terms of FLOPs and improves the accuracy by 3.8% on Epic-Kitchens-100.*

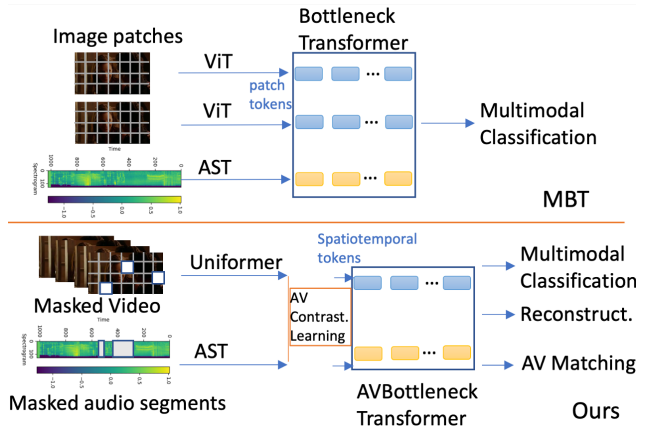

Figure 1. Illustration of training flow of ours (bottom) and previous state-of-the-art multimodal method, MBT [32] (upper).

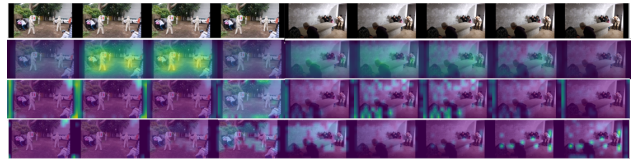

Figure 2. Visualization of one test case of "civil defense siren" in VGGSound [6]. From top to bottom, we show raw video, Grad-CAM [37] of video model [25], AVBottleneck in § 3.2, AVT (ours) with advanced objectives. Video only model and AVBottleneck incorrectly predict it as "people whistling" (focusing people) and "planing timber" (focusing tree) respectively. Understanding multimodal video requires an effective cross-modality learning.

## 1. Introduction

Video understanding [52] has applications including automated event detection, autonomous robots, video ads, video compliance, *etc*. Along with the recent advancement of Transformer [10], several attempts have been made to design video Transformer structures for action recognition [1, 3, 4, 11]. Simply applying a Transformer to 3D video domain is computationally expensive [3]. The Transformer based spatio-temporal learning methods primarily focus on designing efficient variants by factorization along spatial- and temporal-dimensions [3, 4, 48], or employing a multi-scale pyramid structure for a trade-off between the resolution and channel capacity while reducing the memory and computational cost [11, 25, 26, 44, 50, 51, 53].

Utilizing multimodal signals can help extract more representative and complementary feature representations compared to single modality. For instance, audio is extremely useful for recognizing some actions, *e.g.*, dancing, playing musical instruments. As shown in Table 1, audio modality can improve multimodal action recognition for some classes compared to one of the state-of-the-art video-only models. Previous multimodal video Transformers generally employ simple image Transformers. We leverage the latest video Transformer to model complex spatio-temporal features with self-supervised learning, which can fully understand the action from a combination of video and audio input as shown in Fig. 2. An example of multimodal video Transformers, Merlot Reserve [47], conducts audio-vision-language pretraining for holistic multimodal video understanding using an image encoder, word embedding and an audio encoder. MBT [32] constructs multimodal bottleneck tokens to learn video and audio features from image and audio Transformers. The comparison between our method and previous state-of-the-art multimodal method, MBT, is illustrated in Fig. 1, where masked video Transformer, structured masked audio Transformer, and contrastive learning are employed in our method.

In this work, we propose a novel multimodal video transformer, audio-video Transformer (AVT). As illustrated in Fig. 3, AVT effectively employs an audio spectrogram and a video clip as input. Then, a video encoder and an audio encoder are employed to extract video and audio representations respectively. We expect the video encoder to extract complex spatio-temporal representation, which is important to multimodal action recognition. Next, we reduce the cross-modality self-attention complexity through training audio-video bottleneck tokens, which can efficiently learn the cross-modality fusion. To make the model fully understand the semantic content from multimodal signals, we design a novel structured masked audio reconstruction loss where we force the model to reconstruct a whole audio activity segment. Audio-video contrastive loss and audio-video matching loss are designed to reduce the discrepancy between audio and video representations.

The **key contributions** of our work are summarized below:
· We develop audio-video bottleneck Transformer, AVT, which constructs an audio-video bottleneck Transformer to process the embeddings extracted by an audio Transformer and a video Transformer for audio-video recognition.
· We propose a novel masked audio loss which fully exploits the structure of audio spectrogram and predicts a masked whole audio activity segment. Further, contrastive loss is constructed to align the audio and video embedding, and audio-video matching loss is designed to align embeddings after cross-modality fusion.
· Extensive experiments on three public datasets and two in-house datasets consistently demonstrate that, AVT achieves better than its previous state-of-the-art counterparts on Kinetics-Sounds [2] and multimodal approaches on Epic-Kitchens-100 [7] without external training data.

|          | March | Waterfall | Tennis | Laugh | Engi. |
|----------|-------|-----------|--------|-------|-------|
| Uniform. | **83.3** | 46.7 | **100.0** | 14.0 | 39.1 |
| AST      | 81.3  | **60.0**  | 97.6   | **30.2** | 39.1 |

Table 1. Action recognition using video-only, Uniformer [25], and audio-only, AST [16], on the first five categories of VGGSound [6] demonstrate that video and audio are complimentary.

## 2. Related Work

Recently, several works have been conducted using pure-Transformer for spatio-temporal learning [1, 3, 4, 11]. Most of the efforts focus on designing efficient Transformer models to reduce computation and memory consumption. ViViT [3] and TimeSformer [4] study various factorization methods along spatial- and temporal-dimensions. MViT [11, 26] conducts a trade-off between resolution and the number of channels, and constructs a multiscale Transformer to learn a hierarchy from simple dense resolution and fine-grained features to complex coarse features. Multi-view Transformer [44] further employs multiple branches to efficiently learn from various granularities of video views. Uniformer [25] and DualFormer [27] instead modify the internal structures in video Transformer to achieve efficient local-global representation learning by leveraging 3D ConvNets and local-global stratified strategy respectively. VATT [1] conducts unsupervised multi-modality self-supervised pretraining with a pure-Transformer structure for video classification. MBT [32] further constructs multimodal bottleneck tokens to learn multimodal features from an image Transformer and an audio Transformer. Truong *et al*. [39] firstly employ an audiovisual Transformer for speaker localization. Lin *et al*. [29] and Boes and Van hamme [5] employ audiovisual Transformers for event localization and audio classification respectively.

In multimodal self-supervised learning, contrastive self-supervised learning can be used to align multimodal representation from different sources [24, 33]. Li *et al*. [24] firstly propose align before fusion loss using multimodal self-supervised contrastive loss to enhance vision and language representation learning. Yang *et al*. [45] introduce intra-modality contrastive learning into multimodal fusion and obtain a better accuracy. VideoCLIP [43] employs contrastive pre-training for zero-shot video-text understanding. Align and prompt [23] designs entity prompts for effective video-and-language pre-training. OmniMAE [14] shows that one single model masked pretraining can be conducted on images and videos. MAE [12] demonstrates impressive scaling properties for patch masking prediction. Video-MAE [38] proposes customized video tube masking with an extremely high ratio in the pretraining of reconstruction. We propose a novel structured audio segment masking in audio-video representation learning.

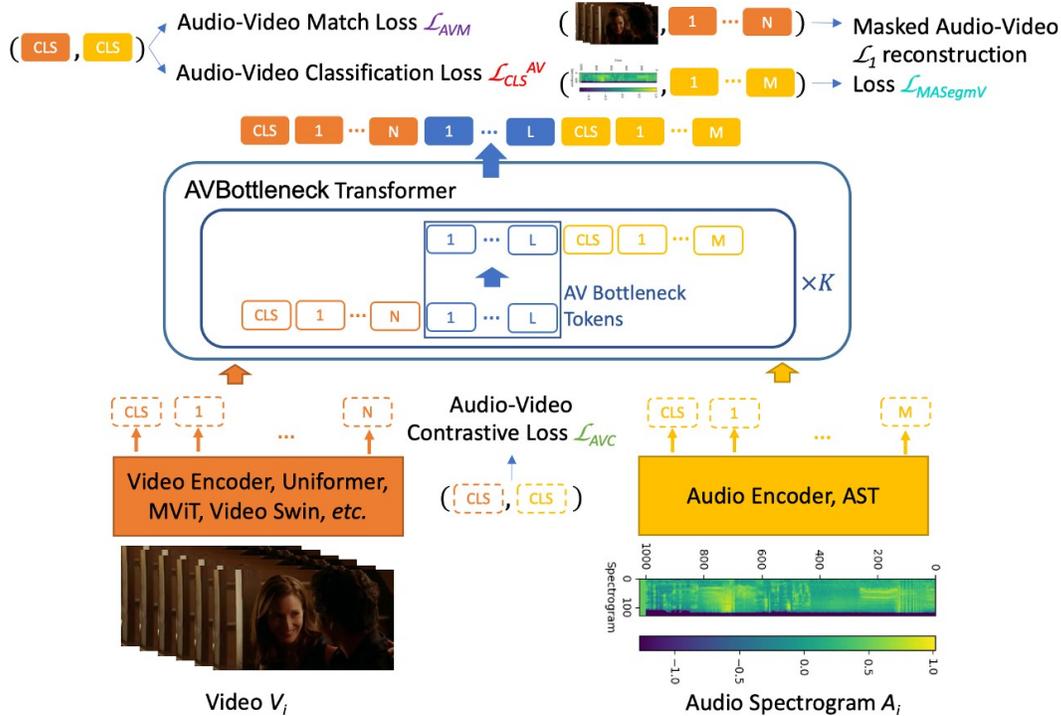

Figure 3. Framework of the audio-video Transformer (best viewed in color), AVT, where multimodal inputs are a video clip $V_i$ and an audio spectrogram $A_i$ from the *i*-th video. Then video encoder and audio encoder are employed to extract video embedding $E^V$ (orange) and audio embedding $E^A$ (yellow) respectively. Next we build audio-video bottleneck tokens, $\{E^F_1, \cdots, E^F_L\}$ (blue), to efficiently learn a cross-modality fusion leveraging self-supervised objectives.

To fully exploit the recent powerful spatio-temporal video Transformer, we construct an audio-video bottleneck Transformer, AVT, employing an audio Transformer and a video Transformer instead of image Transformers in previous multimodal action recognition works.

## 3. Approach

The framework of audio-video Transformer, AVT, is illustrated in Fig. 3, which includes modality encoders, audio-video contrastive loss $\mathcal{L}_{AVC}$, cross-modality bottleneck fusion to reduce computational complexity, audio-video matching loss $\mathcal{L}_{AVM}$ after multimodal fusion, masked audio-video loss $\mathcal{L}_{MASegmV}$ as illustrated in Fig. 4 (b), and multimodal classification loss $\mathcal{L}^{AV}_{CLS}$. The video Transformer extracts discriminative spatio-temporal representation, which is important for multimodal action recognition. Audio-video contrastive loss reduces the distribution discrepancy between audio and video, which benefits the modality fusion and cross-modality learning. Audio-video matching loss and structured masked audio loss considering voice activity structure enable the model to learn high-level semantic representation.

### 3.1. Modality Encoders

Leveraging the recent huge success of audio and video Transformers, we adopt AST [16] as audio modality encoder, and Uniformer [25], MViT [11] or Video Swin Transformer [30] as video modality encoder. Previous image Transformer based work is always constrained by the limited number of input frames, which cannot fully capture the fundamental temporal representation in multimodal action recognition, and we find that a clear accuracy gap exists between powerful image Transformer, CLIP ViT [35], and video Transformers, *i.e.*, MViT [11], Video Swin Transformer [30]. More details can be found in appendix.

Let $V_i$ be the video clip for the *i*-th video, and $A_i$ be the input audio spectrogram. We denote the video encoder as $E^V$ and audio encoder as $E^A$. After the modality encoder, we obtain a video embedding $\{E^V_{CLS}, E^V_1, \cdots, E^V_N\}$ and an audio embedding $\{E^A_{CLS}, E^A_1, \cdots, E^A_M\}$, where $N$ is the total number of tokens in the final layer of video embedding and $M$ is the total number of tokens in the final layer of audio embedding. The cross-entropy loss used to train the video only model is

$$L_{CLS}^V = -\frac{1}{n}\sum_{i=1}^{n}\sum_{c=1}^{C}[y_i(c)\log p_i^V(c)], \quad (1)$$

where $n$ is the batch size in the stochastic gradient descent, $C$ denotes the total number of categories, $y_i$ represents the one-hot ground truth label for the current $i$-th sample, and $p_i^V(c)$ is the video classification probability for label index $c$, which is implemented by a linear layer after $E_{CLS}^V$ with a softmax activation function.

**Audio-video contrastive loss** Multimodal inputs can be considered as different views for the same instance in the contrastive learning. Previous work [24] shows that the image-text contrastive loss yields better accuracy than its counterparts. The cross-modality contrastive learning aligns features from multiple modalities, which benefits the following cross-modality fusion. Thus, we design an audio-video contrastive loss $L_{AVC}$ to align the video and audio representation before cross-modality fusion Transformer

$$\begin{aligned} S_{A,V} &= \exp(g_A(E_{CLS}^A)^T g_V(E_{CLS}^V)), \\ L_{AVC} &= -\mathbb{E}_{(A,V)\in D}[y_{AV}\log\frac{S_{A,V}/\tau}{\sum_{(A,V)\in D}S_{A,V}/\tau}], \end{aligned} \quad (2)$$

where $D$ is the multimodal input that consists of audio $A$ and video $V$, $y_{AV}$ is an indicator that equals to one when the current $A$ and $V$ are from the same sample in the current batch, $\tau$ is a temperature parameter, $g_A$ and $g_V$ are linear embedding layers for audio representation $E^A$ and video representation $E^V$ respectively. The dot product $g_A(\cdot)^T g_V(\cdot)$ measures the similarity of audio and video embedding. Audio-video contrastive learning $L_{AVC}$ penalizes the distribution divergence of audio and video representations for the same sample, which enhances the following cross-modality feature learning.

## 3.2. Cross-Modality Transformer

**AVBottleneck** Previous cross-modality Transformers either simply concatenated multimodal representations [18], or exchanged the key and value matrices between the two modalities [17]. However, due to the huge GPU memory consumption of the existing video Transformer, we construct an audio-video bottleneck Transformer, AVBottleneck, which handles varied lengths of modality tokens efficiently as illustrated in the blue round rectangle of Figure 3. Let $\{E_1^F,\cdots,E_L^F\}$ be the initial multimodal tokens, and $L$ be the number of multimodal tokens. Without loss of generality, we omit the layer number in the denotation. One multimodal bottleneck Transformer block can be formulated as

$$\begin{aligned} E^{VF} &= [E_{CLS}^V, E_1^V, \cdots, E_N^V, E_1^F, \cdots, E_L^F], \\ \tilde{E}^{VF} &= \text{MSA}(\text{LN}(E^{VF})) + E^{VF}, \\ \hat{E}^{VF} &= \text{MLP}(\text{LN}(\tilde{E}^{VF})) + \tilde{E}^{VF}, \end{aligned} \quad (3)$$

$$\begin{aligned} E^{FA} &= [\hat{E}_1^F,\cdots,\hat{E}_L^F, E_{CLS}^A, E_1^A,\cdots, E_M^A], \\ \tilde{E}^{FA} &= \text{MSA}(\text{LN}(E^{FA})) + E^{FA}, \\ \hat{E} &= \text{MLP}(\text{LN}(\tilde{E})) + \tilde{E}, \end{aligned} \quad (4)$$

where the initial multimodal tokens can be updated by averaging the multimodal tokens along all the AVBottleneck blocks. The audio-video bottleneck block can be stacked into $K$ blocks, and the multimodal, video and audio tokens in the following blocks are from the previous block $E_i^F = \hat{E}_i^F$, $E_i^V = \hat{E}_i^V$, and $E_i^A = \hat{E}_i^A$ respectively.

**Computational complexity** AVBottleneck reduces the computing complexity from $O((M+N)^2)$ in merged concatenation based multimodal attention [18] to $O((M+L)^2) + O((N+L)^2) \approx O(M^2) + O(N^2)$, which is the sum of complexity in one block of audio and video Transformers approximately, since $L \ll M, N$. Here, $O(M^2)$ and $O(N^2)$ are the complexities of video and audio Transformers, where $M$ and $N$ are the numbers of tokens in the video and audio Transformers respectively.

**Audio-video matching loss** We design an audio-video matching loss $L_{AVM}$, which can be applied to audio and video embeddings after AVBottleneck and forces the multimodal Transformer to learn high level semantic labels precisely. The discriminative semantic learning can be a cross-entropy loss to determine the current pair is from the same instance or not

$$\begin{aligned} L_{AVM} = -\mathbb{E}_{(A,V)\in D}[&y_{AV}\log p_{AVM}(y_{AV}) + \\ &(1-y_{AV})\log(1-p_{AVM}(y_{AV}))], \end{aligned} \quad (5)$$

where $y_{AV}$ is the same as the audio-video contrastive loss $L_{AVC}$ in Eq. (2), and $p_{AVM}(y_{AV})$ is implemented by concatenating the video and audio embedding $[E_{CLS}^V, E_{CLS}^A]$ followed by a binary classification to determine the sampled audio-video pair $(A, V)$ from the same sample or not in the current batch. Through this cross-modality matching loss $L_{AVM}$, we expect the AVT can effectively learn discriminative features in audio and video cross-modality Transformer. The differences between audio-video matching loss and audio-video contrastive loss mainly are, 1) audio-video contrastive loss is conducted before multimodal fusion to reduce the divergence of the two modality representations, and 2) audio-video matching loss is conducted after multimodal fusion and employs cross-entropy objective to learn discriminative representations.

### 3.2.1 Masked Audio and Video Loss

To learn high-level audio representation, *e.g.*, features for audio activity segments, we further design a masked audio-video loss, $L_{MASegmV}$, in the multimodal Transformer as illustrated in Figure 4, which is much more effective than previous random mask mechanism, $L_{MAV}$, as shown in the ablation study of Table 4. To effectively mask the redundant patches in audio spectrogram, we mask a whole audio activity instead of random masking patches. The audio activity segment can be detected by a second-order smoothing [36] to remove noise, calculating gradient to detect signal change along the time dimension, absolute gradient to detect changes in two temporal directions, and averaging the absolute gradient along the feature dimensions with smoothing to avoid a trivial activity segment. We then choose the top large change points as the transition points to segment different audio activities. In the training, we randomly mask a proportion of the whole complete audio activity segments.

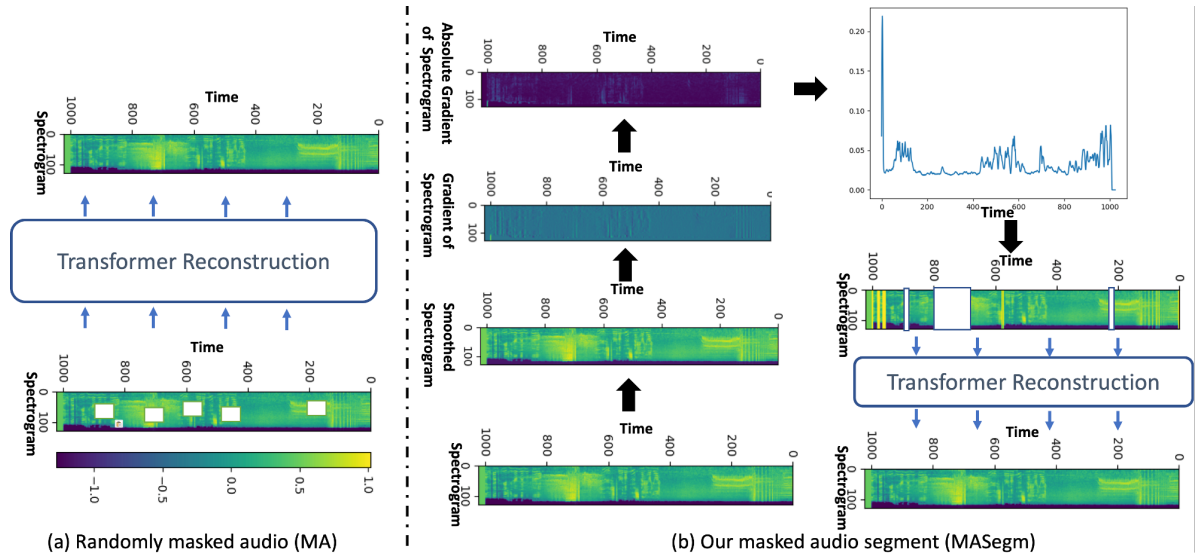

Figure 4. Masked audio models. (a) is a randomly masked audio model, and (b) is our masked audio segment model, which firstly detects the audio activities along the time dimension by smoothing, gradient calculation and averaging absolute gradients along the feature dimension, then applies masks to the whole complete activity along all the features.

$$L_{MASegmV} = \frac{1}{n}\sum_{i=1}^{n} L_1(V_i, \hat{V}_i | \overset{\circ}{V}_i) + L_1(A_i, \hat{A}_i | \overset{\circ}{A}_i), \quad (6)$$

where $\overset{\circ}{V}_i$ is a randomly masked video input, $\overset{\circ}{A}_i$ is a structured (audio complete activity segment) masked audio input, $\hat{V}_i$ and $\hat{A}_i$ are reconstructions from the masked input through the multimodal model, and the decoder can be easily constructed by rearranging the tokens into two or three-dimensional matrix followed by one layer of transposed convolution [46] to match the dimension. If $\overset{\circ}{A}_i$ is a randomly masked audio input, the loss becomes a conventional masked audio and video model $L_{MAV}$. For the randomly masked mechanism, we uniformly choose a proportion of tokens after patch embedding and set these tokens as zero. VideoMAE [38] uses video tube masking and achieves competitive accuracy. Applying the masked activity segments to video Transformer could be a promising topic for future work.

### 3.3. Learning from Multimodal Video

The multimodal classification can be achieved by concatenating the video and audio embedding, $[E^V_{CLS}, E^A_{CLS}]$, and a fully connected layer is constructed to yield the final action classification logits. The supervised multimodal loss can be cross-entropy loss

$$L^{AV}_{CLS} = -\frac{1}{n}\sum_{i=1}^{n}\sum_{c=1}^{C}[y_i'(c)\log p_i^{AV}(c)], \quad (7)$$

where $p_i^{AV}(c)$ is the multimodal classification probability for the $i$-th video and label index $c$.

A hybrid loss considering multimodal video classification and various levels of self-supervised objectives forces the multimodal Transformer to learn effectively from the training data. These self-supervised losses introduce auxiliary objectives to train the multimodal Transformer and act as regularization in the overall supervised learning loss.

$$L = L^{AV}_{CLS} + \lambda_1 L_{AVC} + \lambda_2 L_{AVM} + \lambda_3 L_{MASegmV}, \quad (8)$$

where $\lambda_1$, $\lambda_2$, and $\lambda_3$ are hyperparameters to balance the loss terms in the training. The inference is consistent with the training, and we use the multimodal prediction $p^{AV}$ directly.

## 4. Experimental Results

We experiment with three public audio-video classification datasets – Kinetics-Sounds [2, 19], Epic-Kitchens-100 [7–9], and VGGSound [6].

**Kinetics-Sounds** is a commonly used subset of Kinetics [19], which consists of 10-second videos sampled at 25fps from YouTube. As Kinetics-400 is a dynamic dataset and videos may be removed from YouTube, we follow the dataset collection in Xiao *et al*. [42], and collect 22,914 valid training multimodal videos and 1,585 test videos.

**Epic-Kitchens-100** consists of 90,000 variable length egocentric clips spanning 100 hours capturing daily kitchen activities. The dataset formulates each action into a verb and a noun. We employ two classification heads, one for verb classification and the other one for noun classification, in the AVT. Note that the clips are mainly short-term with average length of 2.6 seconds. Following the standard protocol [7], we report top-1 action-, verb- and noun-accuracies with action accuracy being the primary metric.

**VGGSound** is a large scale action recognition dataset, which consists of about 200K 10-second clips and 309 categories ranging from human actions and sound-emitting objects to human-object interactions. Like other YouTube datasets, *e.g.*, K400 [19], some clips are no longer available. After removing invalid clips, we collect 159,223 valid training multimodal videos and 12,790 valid test videos.

**Implementation details** We investigate two variants of video Transformers, *i.e.*, Uniformer-B 16×4 (#frames×#views) and Uniformer-B 32×4. On Epic-Kitchens-100, we only employ Uniformer-B 16×4 because of the short clip length. We use batch size of 40 and 32 in 8 40 GB NVIDIA A100 GPUs for Uniformer-B with 16 and 32 frames respectively. The numbers of AVBottleneck blocks $K$ and tokens $L$ are all 4, which follows MBT [32]. $\tau$ is fixed as 0.07, and the dimensions of $g_A$ and $g_V$ are set as 256 following Li *et al*. [24]. AdamW [31] is used in the backpropagation and the learning rate is $1 \times 10^{-4}$. The numbers of epochs are 50, 100, 300 on VGGSound, Epic-Kitechens-100 and Kinetics-Sounds respectively. To reduce the effort of tuning hyperparameter of $\lambda_1$, $\lambda_2$ and $\lambda_3$, we validate the hyperparameter one by one as shown in the ablation study while fixing previous validated hyperparameter based on validation set. The $\lambda_1$, $\lambda_2$ and $\lambda_3$ are set as 0.5, 0.1, 0.01. These hyperparameters are generally set to balance the loss values into the same scale, and we did not tune these hyperparameters because of the long training time of each experiment. The number of audio segments is 50, and the masked probability is 4% for both masked audio and video models because the baseline model has already achieved a competitive accuracy. Other hyperparameters follow the recipe in Uniformer [25].

### 4.1. Results

**Comparison to state of the art** AVT surpasses its previous multimodal state-of-the-art counterpart, MBT [32], by 8% and 3.8% on the Kinetics-Sounds and Epic-Kitchens-100 (16 frames) in Table 2 and 3, which demonstrates the video Transformer based multimodal Transformer is better for multimodal action recognition than its image Transformer based counterpart. On VGGSound, AVT achieves slightly better accuracy with the previous state-of-the-art approach, MBT, and is 1.3× more efficient based on the number of FLOPs than MBT [32] from the FLOPs comparison in Table 3 because of the advantage of multiscale mechanism in the video Transformer [25]. Note that MBT uses 10% more training samples than AVT.

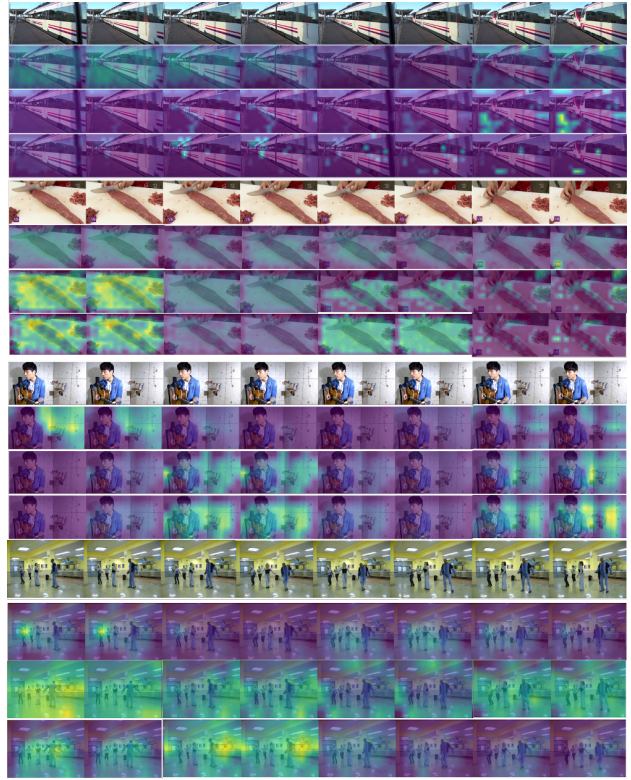

Figure 5. Visualization of four test cases in VGGSound. From top to bottom, we show 8 out of 16 frames of the raw video, Grad-CAM [37] of video only model, AVBottleneck, AVT.

**Ablation study** To verify the effectiveness of each proposed module, we progressively add each loss item to the objective. The ablation results *w.r.t.* audio only, video only, simple averaging audio and video only predictions (Avg), the number of frames, $L_{AVC}$, $L_{AVM}$, random masked model $L_{MAV}$, masked audio segment and video $L_{MASegmV}$, denoted as MASegmV, are concluded in Table 4.

From the Table 4, we find that, (a) AVBottleneck is better than each individual audio and video model. To validate the effectiveness of AVBottleneck, we switch off video input, and denote the model as 'Audio Only'. AVBottleneck outperforms 'Audio Only' by 25.6%, 4.8%, 24.3%, 37.1%, 31.3% on Kinetics-Sounds, VGGSound (16 f), and Epic-

| Models | Modalities | Kinetics-Sounds | | VGGSound | |
|---|---|---|---|---|---|
| | | Top-1 | Top-5 | Top-1 | Top-5 |
| Arandjelovic and Zisserman [2] | A, V | 74.0 | - | - | - |
| AVSlowFast, R101 [42] | A, V | 85.0 | - | - | - |
| Chen *et al.* [6] | A | - | - | 48.8 | 76.5 |
| AudioSlowFast [21] | A | - | - | 50.1 | 77.9 |
| MBT [32] | A | 52.6 | 71.5 | 52.3 | 78.1 |
| MBT [32] | V | 80.7 | 94.9 | 51.2 | 72.6 |
| MBT [32] | A, V | 85.0 | 96.8 | 64.1 | 85.6 |
| AVT | A, V | **93.0** (8%↑) | **99.3** | **64.2** | 85.6 |

Table 2. Comparison to state-of-the-art on Kinetics-Sounds and VGGSound. A: Audio, V: Visual.

| Models | Modalities | Verb | Noun | Action | FLOPs (G) |
|---|---|---|---|---|---|
| Damen *et al.* [7] | A | 42.1 | 21.5 | 14.8 | - |
| AudioSlowFast [21] | A | 46.5 | 22.8 | 15.4 | - |
| TSN [40] | V, F | 60.2 | 46.0 | 33.2 | - |
| TRN [49] | V, F | 65.9 | 45.4 | 35.3 | - |
| TBN [20] | A, V, F | 66.0 | 47.2 | 36.7 | - |
| TSM [28] | V, F | 67.9 | 49.0 | 38.3 | - |
| SlowFast [13] | V | 65.6 | 50.0 | 38.5 | - |
| MBT [32] | A | 44.3 | 22.4 | 13.0 | 131 |
| MBT [32] | V | 62.0 | 56.4 | 40.7 | 140 |
| MBT [32] | A, V | 64.8 | 58.0 | 43.4 | 348 |
| ViViT-L/16×2 [3] | V | 66.4 | 56.8 | 44.0 | 3410 |
| MFormer-HR [34] | V | 67.0 | 58.5 | 44.5 | 959 |
| MeMViT, 16×4 [41] | V | 70.6 | 58.5 | 46.2 | 59 |
| MeMViT [41] (24 frames) | V | 70.6 | 58.5 | 46.2 | 89 |
| MoViNet-A6 [22] (32 frames) | V | 72.2 | 57.3 | 47.7 | 117 |
| Omnivore [15] (32 frames) | V | 69.5 | 61.7 | 49.9 | 492.8 |
| MTV-B [44] (32 frames) | V | 67.8 | 60.5 | 46.7 | 4790 |
| AVT (16 frames) | A, V | 70.4 | 59.3 | 47.2 | 269 |

Table 3. Comparison to previous related work on Epic-Kitchens-100 [7]. F: Optical flow. '-' denotes unavailability from previous work.

Kitchens-100 Verb, Noun, Action. When switching off audio input, we denote the model as 'Video Only'. AVBottleneck outperforms 'Video Only' by 2.2%, 6.6%, 0.9%, 1.0%, 0.8% on the five tasks. (b) Masked audio segment model and other losses are effectiveness. '+AVC', '+AVM' and '+MASegmV' increase top-1 accuracy by 2.0%, 0.4% and 1.1% on VGGSound (16 f). On Kinetics-Sounds, '+MASegmV' surpasses '+AVM' by 0.6% and 1.0% when using 16 and 32 frames. The comprehensive results demonstrate the effectiveness of AVBottleneck and masking audio segment. We also find that models with 32 frames generally perform better than models with 16 frames, which validates that more frames enable more powerful spatio-temporal representation learning in multimodal action recognition.

Previous multimodal approaches, *e.g.*, MBT [32], mostly employ image encoders. On Kinetic-Sounds, even the current state-of-the-art video only model with 16 frames, *i.e.*, Uniformer-B [25], outperforms MBT using an image encoder as the vision backbone in Table 4. This clearly demonstrates the advantage of using a video encoder for vision backbone in the multimodal Transformer. We try our best to reproduce MBT, *i.e.*, Bottleneck with ViT, using 16 frames on VGGSound in Table 5, which demonstrates the effectiveness of temporal modeling in multimodal Transformer and the masked audio segment component.

**Visualizations** We randomly pick four test clips with category names of "train whistling", "chopping food", "playing acoustic guitar", and "people shuffling" from VGGSound test set, and visualize 8 out of 16 frames of raw video, GradCAM [37] of video only model, AVBottleneck, and the fully trained AVT sequentially in Fig. 5. From the first test case (the 1-4th rows), we can find the video only model focuses on general scene and incorrectly predicts this clip as "subway, metro, underground". AVBottleneck incorrectly predicts the clip as "railroad car, train wagon", and the full AVT model with audio and video aligned focuses on different parts of the train and obtains the correct prediction. From the second test case (the 5-8th rows), we find that AVBottleneck in the 7th row cannot capture the knife in the corner and incorrectly predicts the clip as "arc welding". From the third case (the 9-12th rows), we find that the video only model pays attention to the background object and incorrectly predicts the clip as "metronome" and AVT can fully understand the scene. For the last test case (the 13-16th rows), without considering the audio signal, the video only model incorrectly predicts the clip as "tap dancing", which can be easily distinguished from the rhyme of the music.

|  | Kinetics-Sounds | | VGGSound | | Epic-Kitchens-100 | | |
| --- | --- | --- | --- | --- | --- | --- | --- |
| Models | Top-1 | Top-5 | Top-1 | Top-5 | Verb | Noun | Action |
| Audio Only | 66.1 | 88.2 | 54.4 | 76.8 | 45.6 | 22.2 | 15.3 |
| Video Only (16 f) | 89.5 | 98.9 | 52.6 | 75.3 | 69.0 | 58.1 | 45.8 |
| Avg (16 f) | 89.6 | 98.9 | 59.0 | 82.1 | 65.0 | 52.7 | 37.5 |
| AVBottleneck (16 f) | 91.7 | 99.0 | 59.2 | 81.9 | 69.9 | 59.1 | 46.6 |
| +AVC (16 frames) | 92.5 | 99.4 | 61.2 | 82.9 | 70.2 | 58.6 | 46.8 |
| +AVC+AVM (16 f) | 92.4 | 99.5 | 61.6 | 83.4 | 70.3 | 58.9 | 46.9 |
| +AVC+AVM+MAV (16 f) | 92.5 | 99.6 | 61.8 | 83.7 | 70.4 | 58.7 | 46.7 |
| +AVC+AVM+MASegmV (16 f) | 93.0 | 99.2 | 62.7 | 84.9 | **70.4** | **59.3** | **47.2** |
| Video Only (32 f) | 90.7 | 99.1 | 53.2 | 74.8 | | | |
| Avg (32 f) | 90.9 | 98.4 | 58.6 | 82.0 | | | |
| AVBottleneck (32 f) | 91.8 | 99.3 | 58.2 | 80.5 | | | |
| +AVC (32 frames) | 91.8 | 99.5 | 60.7 | 82.2 | | | |
| +AVC+AVM (32 f) | 92.0 | 99.4 | 61.0 | 83.0 | | | |
| +AVC+AVM+MAV (32 f) | 92.4 | 99.3 | 62.4 | 84.8 | | | |
| +AVC+AVM+MASegmV (32 f) | **93.0** | **99.3** | **64.2** | **85.6** | | | |

Table 4. Ablation study on Kinetics-Sounds (left), VGGSound (middle), and Epic-Kitchens-100 (right) [7]. f denotes frames.

| Models | Baseline | w/ MAV | w/ MASegmV |
| --- | --- | --- | --- |
| MBT (Repro.) | 45.5 | 46.9 | 48.1 |
| Ours | 53.7 | 55.6 | 57.1 |

Table 5. Comparison to reproduced MBT [32] with one view on VGGSound based on top-1 accuracy using 16 frames.

We further highlight these difference in the key frames to help better understand the strength of audio-video representation in Fig. 6, which shows 'pumping water' and AVT wrongly classifies as 'toilet flushing' from the water sound and the scene. From the prediction, the model successfully captures the audio signal. In the GradCAM visualization, the model focuses on the cat after the first frame. In the future, more effective audio-video attention, *e.g.*, audio guided attention, can be explored for this case.

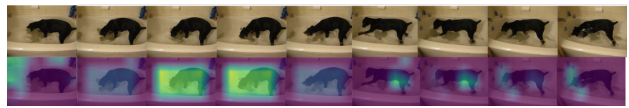

Figure 6. Visualization of 9 out of 16 frames of failed case of 'pumping water'. AVT wrongly classifies as 'toilet flushing' from the water sound and the scene.

## 5. Conclusion

In this work, we have proposed an effective multimodal Transformer, AVT, which leverages advanced video Transformer, audio-video contrastive loss function, audio-video matching loss and a novel masked audio model for multimodal action recognition. These self-supervised objectives penalize different aspects of multimodal Transformer, from reducing the feature divergence before multimodal fusion to forcing to learn high-level semantic representation. AVT surpasses its previous state-of-the-art counterparts by 8% on Kinetics-Sounds. Compared to the previous multimodal Transformer, AVT is 1.3× more efficient in terms of FLOPs and improves the accuracy by 3.8% on Epic-Kitchens-100.

# References

[1] Hassan Akbari, Liangzhe Yuan, Rui Qian, Wei-Hong Chuang, Shih-Fu Chang, Yin Cui, and Boqing Gong. Vatt: Transformers for multimodal self-supervised learning from raw video, audio and text. *Advances in Neural Information Processing Systems*, 34:24206–24221, 2021. 1, 2

[2] Relja Arandjelovic and Andrew Zisserman. Look, listen and learn. In *Proceedings of the IEEE International Conference on Computer Vision*, pages 609–617, 2017. 2, 5, 7

[3] Anurag Arnab, Mostafa Dehghani, Georg Heigold, Chen Sun, Mario Lučić, and Cordelia Schmid. Vivit: A video vision transformer. In *Proceedings of the IEEE/CVF International Conference on Computer Vision*, pages 6836–6846, 2021. 1, 2, 7

[4] Gedas Bertasius, Heng Wang, and Lorenzo Torresani. Is space-time attention all you need for video understanding? In *ICML*, volume 2, page 4, 2021. 1, 2

[5] Wim Boes and Hugo Van hamme. Audiovisual transformer architectures for large-scale classification and synchronization of weakly labeled audio events. In *Proceedings of the 27th ACM International Conference on Multimedia*, pages 1961–1969, 2019. 2

[6] Honglie Chen, Weidi Xie, Andrea Vedaldi, and Andrew Zisserman. Vggsound: A large-scale audio-visual dataset. In *ICASSP 2020-2020 IEEE International Conference on Acoustics, Speech and Signal Processing (ICASSP)*, pages 721–725. IEEE, 2020. 1, 2, 5, 7

[7] Dima Damen, Hazel Doughty, Giovanni Maria Farinella, , Antonino Furnari, Jian Ma, Evangelos Kazakos, Davide Moltisanti, Jonathan Munro, Toby Perrett, Will Price, and Michael Wray. Rescaling egocentric vision: Collection, pipeline and challenges for epic-kitchens-100. *International Journal of Computer Vision (IJCV)*, 2021. 2, 5, 6, 7, 8

[8] Dima Damen, Hazel Doughty, Giovanni Maria Farinella, Sanja Fidler, Antonino Furnari, Evangelos Kazakos, Davide Moltisanti, Jonathan Munro, Toby Perrett, Will Price, and Michael Wray. Scaling egocentric vision: The epic-kitchens dataset. In *European Conference on Computer Vision (ECCV)*, 2018. 5

[9] Dima Damen, Hazel Doughty, Giovanni Maria Farinella, Sanja Fidler, Antonino Furnari, Evangelos Kazakos, Davide Moltisanti, Jonathan Munro, Toby Perrett, Will Price, and Michael Wray. The epic-kitchens dataset: Collection, challenges and baselines. *IEEE Transactions on Pattern Analysis and Machine Intelligence (TPAMI)*, 43(11):4125–4141, 2021. 5

[10] Alexey Dosovitskiy, Lucas Beyer, Alexander Kolesnikov, Dirk Weissenborn, Xiaohua Zhai, Thomas Unterthiner, Mostafa Dehghani, Matthias Minderer, Georg Heigold, Sylvain Gelly, et al. An image is worth 16x16 words: Transformers for image recognition at scale. In *International Conference on Learning Representations*, 2021. 1

[11] Haoqi Fan, Bo Xiong, Karttikeya Mangalam, Yanghao Li, Zhicheng Yan, Jitendra Malik, and Christoph Feichtenhofer. Multiscale vision transformers. In *Proceedings of the IEEE/CVF International Conference on Computer Vision*, pages 6824–6835, 2021. 1, 2, 3

[12] Christoph Feichtenhofer, Haoqi Fan, Yanghao Li, and Kaiming He. Masked autoencoders as spatiotemporal learners. In *Advances in Neural Information Processing Systems*, 2022. 2

[13] Christoph Feichtenhofer, Haoqi Fan, Jitendra Malik, and Kaiming He. Slowfast networks for video recognition. In *Proceedings of the IEEE/CVF International Conference on Computer Vision*, pages 6202–6211, 2019. 7

[14] Rohit Girdhar, Alaaeldin El-Nouby, Mannat Singh, Kalyan Vasudev Alwala, Armand Joulin, and Ishan Misra. Omnimae: Single model masked pretraining on images and videos. *arXiv preprint arXiv:2206.08356*, 2022. 2

[15] Rohit Girdhar, Mannat Singh, Nikhila Ravi, Laurens van der Maaten, Armand Joulin, and Ishan Misra. Omnivore: A single model for many visual modalities. In *Proceedings of the IEEE/CVF Conference on Computer Vision and Pattern Recognition*, pages 16102–16112, 2022. 7

[16] Yuan Gong, Yu-An Chung, and James Glass. Ast: Audio spectrogram transformer. In *Proceedings of Interspeech*, 2021. 2, 3

[17] Lisa Anne Hendricks, John Mellor, Rosalia Schneider, Jean-Baptiste Alayrac, and Aida Nematzadeh. Decoupling the role of data, attention, and losses in multimodal transformers. *Transactions of the Association for Computational Linguistics*, 9:570–585, 2021. 4

[18] Andrew Jaegle, Felix Gimeno, Andy Brock, Oriol Vinyals, Andrew Zisserman, and Joao Carreira. Perceiver: General perception with iterative attention. In *International Conference on Machine Learning*, pages 4651–4664. PMLR, 2021. 4

[19] Will Kay, Joao Carreira, Karen Simonyan, Brian Zhang, Chloe Hillier, Sudheendra Vijayanarasimhan, Fabio Viola, Tim Green, Trevor Back, Paul Natsev, et al. The kinetics human action video dataset. *arXiv preprint arXiv:1705.06950*, 2017. 5, 6

[20] Evangelos Kazakos, Arsha Nagrani, Andrew Zisserman, and Dima Damen. Epic-fusion: Audio-visual temporal binding for egocentric action recognition. In *Proceedings of the IEEE/CVF International Conference on Computer Vision*, pages 5492–5501, 2019. 7

[21] Evangelos Kazakos, Arsha Nagrani, Andrew Zisserman, and Dima Damen. Slow-fast auditory streams for audio recognition. In *ICASSP 2021-2021 IEEE International Conference on Acoustics, Speech and Signal Processing (ICASSP)*, pages 855–859. IEEE, 2021. 7

[22] Dan Kondratyuk, Liangzhe Yuan, Yandong Li, Li Zhang, Mingxing Tan, Matthew Brown, and Boqing Gong. Movinets: Mobile video networks for efficient video recognition. In *Proceedings of the IEEE/CVF Conference on Computer Vision and Pattern Recognition*, pages 16020–16030, 2021. 7

[23] Dongxu Li, Junnan Li, Hongdong Li, Juan Carlos Niebles, and Steven CH Hoi. Align and prompt: Video-and-language pre-training with entity prompts. In *Proceedings of the IEEE/CVF Conference on Computer Vision and Pattern Recognition*, pages 4953–4963, 2022. 2

[24] Junnan Li, Ramprasaath Selvaraju, Akhilesh Gotmare, Shafiq Joty, Caiming Xiong, and Steven Chu Hong Hoi.

Align before fuse: Vision and language representation learning with momentum distillation. *Advances in Neural Information Processing Systems*, 34, 2021. 2, 4, 6

[25] Kunchang Li, Yali Wang, Peng Gao, Guanglu Song, Yu Liu, Hongsheng Li, and Yu Qiao. Uniformer: Unified transformer for efficient spatiotemporal representation learning. In *International Conference on Learning Representations*, 2022. 1, 2, 3, 6, 7

[26] Yanghao Li, Chao-Yuan Wu, Haoqi Fan, Karttikeya Mangalam, Bo Xiong, Jitendra Malik, and Christoph Feichtenhofer. Improved multiscale vision transformers for classification and detection. In *Proceedings of the IEEE Conference on Computer Vision and Pattern Recognition*, 2022. 1, 2

[27] Yuxuan Liang, Pan Zhou, Roger Zimmermann, and Shuicheng Yan. Dualformer: Local-global stratified transformer for efficient video recognition. In *European Conference on Computer Vision*, 2022. 2

[28] Ji Lin, Chuang Gan, and Song Han. Tsm: Temporal shift module for efficient video understanding. In *Proceedings of the IEEE/CVF International Conference on Computer Vision*, pages 7083–7093, 2019. 7

[29] Yan-Bo Lin and Yu-Chiang Frank Wang. Audiovisual transformer with instance attention for audio-visual event localization. In *Proceedings of the Asian Conference on Computer Vision*, 2020. 2

[30] Ze Liu, Jia Ning, Yue Cao, Yixuan Wei, Zheng Zhang, Stephen Lin, and Han Hu. Video swin transformer. In *Proceedings of the IEEE/CVF Conference on Computer Vision and Pattern Recognition*, pages 3202–3211, 2022. 3

[31] Ilya Loshchilov and Frank Hutter. Decoupled weight decay regularization. In *International Conference on Learning Representations*, 2018. 6

[32] Arsha Nagrani, Shan Yang, Anurag Arnab, Aren Jansen, Cordelia Schmid, and Chen Sun. Attention bottlenecks for multimodal fusion. *Advances in Neural Information Processing Systems*, 34, 2021. 1, 2, 6, 7, 8

[33] Aaron van den Oord, Yazhe Li, and Oriol Vinyals. Representation learning with contrastive predictive coding. *arXiv preprint arXiv:1807.03748*, 2018. 2

[34] Mandela Patrick, Dylan Campbell, Yuki Asano, Ishan Misra, Florian Metze, Christoph Feichtenhofer, Andrea Vedaldi, and Joãol F Henriques. Keeping your eye on the ball: Trajectory attention in video transformers. *Advances in Neural Information Processing Systems*, 34:12493–12506, 2021. 7

[35] Alec Radford, Jong Wook Kim, Chris Hallacy, Aditya Ramesh, Gabriel Goh, Sandhini Agarwal, Girish Sastry, Amanda Askell, Pamela Mishkin, Jack Clark, et al. Learning transferable visual models from natural language supervision. In *International Conference on Machine Learning*, pages 8748–8763. PMLR, 2021. 3

[36] Abraham Savitzky and Marcel JE Golay. Smoothing and differentiation of data by simplified least squares procedures. *Analytical Chemistry*, 36(8):1627–1639, 1964. 5

[37] Ramprasaath R Selvaraju, Michael Cogswell, Abhishek Das, Ramakrishna Vedantam, Devi Parikh, and Dhruv Batra. Grad-cam: Visual explanations from deep networks via gradient-based localization. In *Proceedings of the IEEE International Conference on Computer Vision*, pages 618–626, 2017. 1, 6, 8

[38] Zhan Tong, Yibing Song, Jue Wang, and Limin Wang. Videomae: Masked autoencoders are data-efficient learners for self-supervised video pre-training. In *Advances in Neural Information Processing Systems*, 2022. 2, 5

[39] Thanh-Dat Truong, Chi Nhan Duong, Hoang Anh Pham, Bhiksha Raj, Ngan Le, Khoa Luu, et al. The right to talk: An audio-visual transformer approach. In *Proceedings of the IEEE/CVF International Conference on Computer Vision*, pages 1105–1114, 2021. 2

[40] Limin Wang, Yuanjun Xiong, Zhe Wang, Yu Qiao, Dahua Lin, Xiaoou Tang, and Luc Van Gool. Temporal segment networks: Towards good practices for deep action recognition. In *European Conference on Computer Vision*, pages 20–36. Springer, 2016. 7

[41] Chao-Yuan Wu, Yanghao Li, Karttikeya Mangalam, Haoqi Fan, Bo Xiong, Jitendra Malik, and Christoph Feichtenhofer. Memvit: Memory-augmented multiscale vision transformer for efficient long-term video recognition. In *Proceedings of the IEEE/CVF Conference on Computer Vision and Pattern Recognition*, pages 13587–13597, 2022. 7

[42] Fanyi Xiao, Yong Jae Lee, Kristen Grauman, Jitendra Malik, and Christoph Feichtenhofer. Audiovisual slowfast networks for video recognition. *arXiv preprint arXiv:2001.08740*, 2020. 6, 7

[43] Hu Xu, Gargi Ghosh, Po-Yao Huang, Dmytro Okhonko, Armen Aghajanyan, Florian Metze, Luke Zettlemoyer, and Christoph Feichtenhofer. Videoclip: Contrastive pre-training for zero-shot video-text understanding. In *Proceedings of the 2021 Conference on Empirical Methods in Natural Language Processing*, pages 6787–6800, 2021. 2

[44] Shen Yan, Xuehan Xiong, Anurag Arnab, Zhichao Lu, Mi Zhang, Chen Sun, and Cordelia Schmid. Multiview transformers for video recognition. In *Proceedings of the IEEE Conference on Computer Vision and Pattern Recognition*, 2022. 1, 2, 7

[45] Jinyu Yang, Jiali Duan, Son Tran, Yi Xu, Sampath Chanda, Liqun Chen, Belinda Zeng, Trishul Chilimbi, and Junzhou Huang. Vision-language pre-training with triple contrastive learning. In *Proceedings of the IEEE/CVF Conference on Computer Vision and Pattern Recognition*, pages 15671–15680, 2022. 2

[46] Matthew D Zeiler, Dilip Krishnan, Graham W Taylor, and Rob Fergus. Deconvolutional networks. In *2010 IEEE Computer Society Conference on Computer Vision and Pattern Recognition*, pages 2528–2535. IEEE, 2010. 5

[47] Rowan Zellers, Jiasen Lu, Ximing Lu, Youngjae Yu, Yanpeng Zhao, Mohammadreza Salehi, Aditya Kusupati, Jack Hessel, Ali Farhadi, and Yejin Choi. Merlot reserve: Neural script knowledge through vision and language and sound. In *Proceedings of the IEEE/CVF Conference on Computer Vision and Pattern Recognition*, pages 16375–16387, 2022. 2

[48] Xuefan Zha, Wentao Zhu, Lv Xun, Sen Yang, and Ji Liu. Shifted chunk transformer for spatio-temporal representational learning. In *Advances in Neural Information Processing Systems*, volume 34, 2021. 1

[49] Bolei Zhou, Alex Andonian, Aude Oliva, and Antonio Torralba. Temporal relational reasoning in videos. In *Proceedings of the European Conference on Computer Vision (ECCV)*, pages 803–818, 2018. 7

[50] Wentao Zhu, Mohamed Omar. Multiscale Audio Spectrogram Transformer for Efficient Audio Classification. In ICASSP, 2023.

[51] Wentao Zhu. Efficient Multiscale Multimodal Bottleneck Transformer for Audio-Video Classification. arXiv, 2024.

[52] Wentao Zhu, Cuiling Lan, Junliang Xing, Wenjun Zeng, Yanghao Li, Li Shen, Xiaohui Xie. Co-occurrence feature learning for skeleton based action recognition using regularized deep LSTM networks. AAAI, 2016.

[53] Xuefan Zha, Wentao Zhu, Tingxun Lv, Sen Yang, and Ji Liu. Shifted chunk transformer for spatio-temporal representational learning. In *Advances in Neural Information Processing Systems, 2021.*